\documentclass[sigconf]{acmart}
\AtBeginDocument{%
  }

\usepackage{tikz}
\usetikzlibrary{arrows.meta, positioning, fit, shapes.misc, shapes.geometric, shadows, backgrounds, positioning}
\usepackage{subcaption}

\usepackage{amsmath}
\usepackage[english]{babel}  

\begin{document}
\copyrightyear{2026}
\acmYear{2026}
\setcopyright{rightsretained}
\setcctype{by}
\acmConference[CAIN '26]{2026 IEEE/ACM 5th International Conference on AI Engineering - Software Engineering for AI}{April 12--13, 2026}{Rio de Janeiro, Brazil}
\acmBooktitle{2026 IEEE/ACM 5th International Conference on AI Engineering - Software Engineering for AI (CAIN '26), April 12--13, 2026, Rio de Janeiro, Brazil}
\acmPrice{}
\acmDOI{10.1145/3793653.3793774}
\acmISBN{979-8-4007-2475-6/2026/04}
\title{Engineering AI Agents for Clinical Workflows: A Case Study in Architecture, MLOps, and Governance}

\author{Cláudio Lúcio do Val Lopes}
\email{claudio.lucio@a3data.com.br}
\orcid{0000-0003-1655-2283}
\affiliation{%
  \institution{A3Data}
  \city{BH}
  \state{MG}
  \country{Brazil}
}
\affiliation{%
  \institution{CEFET-MG}
  \city{BH}
  \state{MG}
  \country{Brazil}
}

\author{João Marcus Pitta}
\email{joao.marcus@a3data.com.br }
\affiliation{%
  \institution{A3Data}
  \city{BH}
  \state{MG}
  \country{Brazil}
}

\author{Fabiano Belém}
\email{fabiano.belem@a3data.com.br}
\affiliation{%
  \institution{A3Data}
  \city{BH}
  \state{MG}
  \country{Brazil}
}

\author{Gildson Alves}
\email{gildson.alves@a3data.com.br}
\affiliation{%
  \institution{A3Data}
  \city{BH}
  \state{MG}
  \country{Brazil}
}

\author{Flávio Vinícius Cruzeiro Martins}
\email{flaviocruzeiro@cefetmg.br}
\affiliation{%
  \institution{CEFET-MG}
  \city{BH}
  \state{MG}
  \country{Brazil}
}

\renewcommand{\shortauthors}{Cláudio et al.}

%%
%% sis abstract is a short summary of the work to be presented in the

\begin{abstract}
The integration of Artificial Intelligence (AI) into clinical settings presents a software engineering challenge, demanding a shift from isolated models to robust, governable, and reliable systems. However, brittle, prototype-derived architectures often plague industrial applications and a lack of systemic oversight, creating a ``responsibility vacuum'' where safety and accountability are compromised. This paper presents an industry case study of the ``Maria'' platform, a production-grade AI system in primary healthcare that addresses this gap.

Our central hypothesis is that trustworthy clinical AI is achieved through the holistic integration of four foundational engineering pillars. We present a synergistic architecture that combines Clean Architecture for maintainability with an Event-driven architecture for resilience and auditability. We introduce the Agent as the primary unit of modularity, each possessing its own autonomous MLOps lifecycle. Finally, we show how a Human-in-the-Loop governance model is technically integrated not merely as a safety check, but as a critical, event-driven data source for continuous improvement. We present the platform as a reference architecture, offering practical lessons for engineers building maintainable, scalable, and accountable AI-enabled systems in high-stakes domains.

\end{abstract}

\begin{CCSXML}
<ccs2012>
<concept>
<concept_id>10011007.10011006.10011066.10011070</concept_id>
<concept_desc>Software and its engineering~Application specific development environments</concept_desc>
<concept_significance>500</concept_significance>
</concept>
</ccs2012>
\end{CCSXML}

\ccsdesc[500]{Software and its engineering~Application specific development environments}

\keywords{AI Engineering, Software Architecture, MLOps, AI Governance, Human-in-the-Loop (HITL), AI Agents, Healthcare Systems}


\maketitle

\section{Introduction}
\label{sec:introduction}

The successful integration of Artificial Intelligence (AI) into clinical settings represents one of the most significant engineering challenges of our time \cite{Lekadir2023FutureAI, app15031344}. This task demands a critical transition from developing isolated, high-performing models in laboratory environments to engineering robust, reliable, and governable systems that can function safely within the complex socio-technical fabric of healthcare \cite{Anon2024ToolsToAgents}.

However, the common practice of evolving data science prototypes, often originating in Jupyter notebooks, into production systems frequently results in monolithic, tightly coupled applications that are difficult to debug, maintain, and extend. This anti-pattern arises because the goals of exploratory data science (rapid iteration and discovery) are misaligned with the goals of production engineering (robustness, testability, and maintainability). These prototypes often suffer from tangled code, where data cleaning, feature engineering, and model logic are intertwined, and rely on a hidden global state that makes their behavior difficult to reproduce. When this code is copied directly into a production service, it creates a system with high technical debt and no clear separation of concerns, rendering it unsafe for a high-stakes clinical environment.

This accumulation of hidden technical debt is untenable in clinical systems where reliability and safety are non-negotiable. Furthermore, a lack of ongoing monitoring and oversight often creates a ``responsibility vacuum'' in AI governance, posing a substantial risk of patient harm as models inevitably degrade or encounter unforeseen edge cases in production \cite{Anon2025ResponsibilityVacuum}. The field lacks comprehensive industrial case studies in the healthcare domain that demonstrate how to bridge this gap from a rigorous software engineering perspective \cite{Warrier2025RealTimeAI}.

This paper presents an industry case study of the ``Maria'' platform \cite{MariaSaude:2025}, a primary healthcare system in production that utilizes AI agents to optimize clinical workflows. The central hypothesis of our work is not the novelty of our Natural Language Processing (NLP) models, the use of generative AI, or Agentic approach \cite{Anon2024ToolsToAgents}, but rather the holistic and principled software engineering integration of four foundational pillars:
\begin{enumerate}
    \item \textbf{Architecture:} A synergistic combination of \textit{Clean Architecture} to protect core domain logic and an \textit{Event-Driven Architecture (EDA)} to enable real-time responsiveness and auditability;
    \item \textbf{Agent-Based Design:} The conceptualization of AI components (Pre- and Post-medical appointment agents) as autonomous agents, each with its own distinct lifecycle and responsibilities;
    \item \textbf{MLOps:} The implementation of robust, autonomous MLOps pipelines for each agent, focusing on versioning, contextual performance monitoring, and observability;
    \item \textbf{Governance:} The application of a \textit{Human-in-the-Loop (HITL)} model for supervised medical validation to ensure safety, accountability, and clinical trust.
\end{enumerate}

This case study is guided by research questions focused on these engineering challenges:

\begin{itemize}
    \item \textbf{RQ1 (Architecture):} How can Clean Architecture and Event-Driven Architecture be synergistically combined to build a resilient, maintainable, and auditable clinical AI system? 
    \item \textbf{RQ2 (Design \& MLOps):} How does conceptualizing AI models as autonomous agents, each with a dedicated MLOps lifecycle, improve system modularity, governance, and maintainability? 
    \item \textbf{RQ3 (Governance):} How can a Human-in-the-Loop (HITL) governance model be technically integrated with MLOps pipelines to ensure safety and accountability in a live clinical workflow?
\end{itemize}

Our primary contributions to the practice of AI engineering are a case study of a synergistic architectural pattern (Clean + EDA) for high-stakes AI systems, demonstrating how it protects core business logic while enabling decoupled, responsive, and inherently auditable components. Secondly, a practical design paradigm that frames the ``Agent'' as the fundamental unit of modularity, deployment, and governance in AI-enabled systems, where each agent possesses an autonomous and observable MLOps lifecycle. 
Additionally, we discuss an integrated MLOps-Governance framework, showing how a robust MLOps pipeline 
is an essential technical prerequisite for a meaningful HITL validation process, it also provides the traceability and context required for safe clinical oversight. Finally, we bring lessons learned from production deployment, detailing how this integrated approach fills the governance gap and serves as a reference architecture for other practitioners.

The remainder of this paper is structured as follows: Section \ref{sec:related} reviews related work in AI system architecture, healthcare agents, and MLOps. Section \ref{sec:system} presents the ``Maria''  system as an AI engineering case study, detailing its system architecture, agent-based design, MLOps lifecycle, and its AI governance. Section \ref{sec:discussion} discusses our results and the lessons learned from the field. Finally, Section \ref{sec:conclusion} concludes the paper and outlines future work.

\section{Related work} \label{sec:related}
This paper integrates four distinct but interdependent domains of AI engineering. Our contribution lies not in a single domain, but in its holistic integration to solve a real-world clinical challenge. We structure our review around these four pillars.

\subsection{Architectural patterns for AI-enabled systems}
\label{sec:related:architecture}

The challenge of moving AI from prototype to production has revealed the limitations of traditional, monolithic software design. The common practice of evolving notebook-based prototypes often leads to tightly coupled, brittle systems with high technical debt \cite{app15031344}. In response, the field has begun to adopt more principled software architectures.

Clean Architecture, as formalized by Martin, provides a robust solution by separating concerns into concentric layers \cite{10.5555/3175742}. Its central dependency rule ensures that core domain and application logic (the ``Entities'' and ``Use Cases'') are independent of volatile outer layers like frameworks, databases, and external APIs. This framework independence is critical for the long-term maintainability and testability of high-stakes systems, as it protects stable clinical business rules from technological churn.

Simultaneously, Event-Driven Architecture (EDA) has emerged as a dominant pattern for building scalable, responsive, and resilient systems. At the heart of modern EDA is the publish-subscribe model \cite{10.1145/857076.857078}. In this pattern, components communicate asynchronously by producing events (as publishers) and consuming events (as subscribers), without any direct knowledge of each other. This creates powerful decoupling in space, time, and synchronization \cite{Surantha2022, Wang2013}. This pattern is exceptionally well-suited to the fragmented healthcare ecosystem, enabling real-time responsiveness to clinical events (e.g., patient check-in, lab result arrival) and the integration of disparate systems like Electronic Health Records(EHRs) and pharmacy modules \cite{Sucipto2025, Reolid2023}.

While the literature discusses these patterns, they are often treated in isolation. Our work contributes with a case study on their {synergistic combination}: using Clean Architecture to provide a resilient, testable core for each component, and EDA to orchestrate these components into a responsive, auditable, and decoupled whole.

\subsection{MLOps for production AI in healthcare}
\label{sec:related:mlops}

Machine Learning Operations (MLOps) applies DevOps principles to the machine learning lifecycle, addressing challenges in reproducibility, automated testing, continuous integration (CI), continuous delivery (CD), and monitoring \cite{10081336}. The goal is to streamline and automate the process of bringing models into production and maintaining their performance over time.

In regulated environments like healthcare, these practices, sometimes termed ``MedMLOps'' are prerequisites for safety and compliance. The challenges are amplified by the need to handle sensitive data (HIPAA/GDPR \footnote{
HIPAA is a US federal law that governs the privacy and security of Personal Health
Information (PHI) in the US.
GDPR extends to General Data Protection Regulation, it is a legal framework that sets guidelines for
the collection and processing of personal information from individuals who live in the
European Union.}). To address this, robust MLOps frameworks must enforce strict data minimization strategies upstream of the Ingestion API, ensuring that only essential clinical context enters the pipeline while reducing the exposure of Protected Health Information (PHI). Furthermore, systems must detect model drift as clinical practices or patient populations change, and integrate with legacy IT infrastructure \cite{Anon2025MedMLOps}. MLOps also ensures that every prediction can be traced to a specific, versioned model with known performance characteristics, which is fundamental for building and maintaining clinical trust. Our work builds on this by demonstrating a granular MLOps implementation at the ``Agent'' level, providing an autonomous, observable lifecycle for each AI component.

\subsection{AI Agents in healthcare}
\label{sec:related:agents}

The paradigm for AI in medicine is undergoing a significant shift, moving beyond viewing AI as a set of disconnected {tools} for specific, isolated tasks (e.g., classification, regression, clustering) and toward designing intelligent, autonomous {Agents}. An agent, in this context, is an entity that can perceive its environment, reason about its context, and take actions to achieve a specific goal\cite{li2025agenthospitalsimulacrumhospital}. This approach is particularly relevant in primary care, which faces an immense burden from administrative and documentation tasks.

Consequently, AI agents are being actively explored to automate and augment these workflows. The literature highlights several key applications. Reducing Administrative Burden is the most prominent use case. Agents are being developed to function as ambient scribes that listen to patient-clinician conversations to automatically generate clinical notes, thereby reducing documentation time and professional burnout \cite{10.1001/jamanetworkopen.2025.34976}. Other tasks include optimizing billing, reducing insurance-related expenses, and improving quality assurance \cite{Lin2019}.

In supporting clinical workflows, agents are also being designed to synthesize patient histories, support clinical decision-making \cite{LAVOIEGAGNE20253270, 10.1001/jamainternmed.2023.7965}, and manage administrative tasks, ultimately aiming to improve care for both patients and clinicians\cite{Terrye90}.

However, much of the existing literature describes these applications as single-point solutions that address only one isolated task, e.g., only documentation, or only history summarization. Our ``Maria'' platform offers a more mature, workflow-centric design. We deploy a dual-agent architecture that ``bookends'' the entire clinical encounter (Pre- and Post-medical appointments). This design reframes the AI component from `a model to be managed' to `an autonomous agent' that acts as a true collaborator in the clinical process.

\subsection{Governance and Human-in-the-Loop (HITL)}
\label{sec:related:governance}

Given the high-stakes nature of medicine, fully autonomous AI systems are often inappropriate and unsafe \cite{10.1001/jamainternmed.2023.7965}. This has led to the emergence of a ``responsibility vacuum'' where the ongoing monitoring and oversight of AI in production are poorly defined \cite{Anon2025ResponsibilityVacuum}. Human-in-the-Loop (HITL) has become the foundational governance strategy to address this gap \cite{10.1145/3377325.3377489}. HITL strategically inserts a qualified human, in our case, a clinician at a critical decision point to review, validate, correct, or reject the AI's output before it has a clinical impact \cite{info:doi/10.2196/69678}.

HITL is more than a user-interface pattern; it is a core mechanism for accountability. However, the literature often discusses HITL in isolation from the underlying engineering systems. A key contribution of our work is to demonstrate that a mature MLOps pipeline is the essential technical prerequisite for a meaningful and scalable HITL governance model \cite{Anon2025MLOpsMaturity}. A clinician cannot provide effective oversight if they are validating the output of a black box. Our system shows how MLOps delivers the essential context for governance, linking every AI suggestion to a specific version of the model, its training data, and its performance metrics, thus transforming HITL into a truly auditable and accountable clinical safety net.

\subsection{Identifying the gap}
Although the literature provides valuable information about these individual pillars, there is a discernible gap in comprehensive industrial case studies. Few papers only partially discuss how to holistically integrate Clean Architecture, EDA, granular agent-based MLOps, or a technically-grounded HITL governance model into a single, functional, and production-grade clinical system \cite{Huang2024, Mallardi2024MLOps, Anon2025ResponsibilityVacuum, Anon2024ToolsToAgents, Surantha2022}. This paper aims to fill this gap by presenting such a full case study, providing a reference architecture, and practical lessons for other practitioners in the field.

\section{The ``Maria'' system: An AI engineering case study} \label{sec:system}

The ``Maria'' platform \cite{MariaSaude:2025} is a production-grade clinical support system designed to address the engineering challenges of building reliable, maintainable, and governable AI. Its design is a direct answer to the research questions posed in Section \ref{sec:introduction} and is built upon the four pillars of architecture, agent-based design, MLOps, and governance. This section details the ``Maria''' system, beginning with its foundational architecture.

\subsection{A synergistic achitecture for resilience and maintainability (RQ1)}
\label{sec:system:architecture}

\begin{figure*}[t]
\centering
\resizebox{\textwidth}{!}{%
\begin{tikzpicture}[
    node distance=1.2cm and 2.0cm,
    % --- Define Styles ---
    comp/.style={
        draw, 
        rectangle, 
        thick,
        fill=white,
        minimum height=1.7cm, 
        text width=4cm,
        text centered,
        font=\sffamily\large  % <-- FONT INCREASED
    },
    container/.style={
        draw,
        rectangle,
        thick,
        inner sep=1cm,
        label={[font=\sffamily\bfseries\large]north:#1}, % <-- FONT INCREASED
        fill=none
    },
    arrow/.style={
        draw, 
        thick, 
        -Stealth
    },
    edge_label/.style={
        fill=gray!20,
        font=\sffamily\small, % <-- FONT INCREASED
        inner sep=2pt,
        pos=0.5
    }
]

% --- Node Placement (Unchanged) ---
\node[comp, label={[font=\sffamily\bfseries\large]north:Client}] (client) {User Input \\ \footnotesize (Text \textbullet{} Voice \textbullet{} Form)}; %<-- Label font increased
\node[comp, right=of client, label={[font=\sffamily\bfseries\large]north:API}] (api) {Ingestion API \\ \footnotesize (HTTP / WS)}; %<-- Label font increased
\node[comp, right=of api] (intent) {Intent Classifier \\ \footnotesize (LLM + Cache)};
\node[comp, right=of intent] (agent) {Agent Layer \\ \footnotesize (Pre \& Post-medical appointment)};
\node[comp, above=of agent] (usecases) {Use Cases / Policies \\ \footnotesize (Clean Architecture Core)};
\node[comp, right=2.5cm of agent] (eventbus) {Event Bus \\ \footnotesize (EventBridge / SNS)};
\node[comp, below=of eventbus] (storage) {Storage \\ \footnotesize (Aurora \textbullet{} S3 \textbullet{} DynamoDB)};
\node[comp, below=of storage] (llm) {LLM Providers \\ \footnotesize (OpenAI / Bedrock)};
\node[comp, right=of eventbus] (hitl) {HITL Workflow \\ \footnotesize (Step Functions + Clinician Review)};
\node[comp, right=of storage] (obs) {Observability \\ \footnotesize (Logs \textbullet{} Metrics \textbullet{} Traces)};
\node[comp, below=of llm, yshift=-1cm] (response) {Response Delivery \\ \footnotesize (App / Portal / WhatsApp)};

% --- Fit Containers (Unchanged) ---
\node[container, 
      fit=(intent) (agent) (usecases), 
      label={[font=\sffamily\bfseries\large]north:Core} %<-- Label font increased
      ] (core_box) {};
      
\node[container, 
      fit=(eventbus) (storage) (llm) (hitl) (obs), 
      label={[font=\sffamily\bfseries\large]north:Infrastructure} %<-- Label font increased
      ] (infra_box) {};

% --- Draw Arrows & Edge Labels (Unchanged) ---
\draw[arrow] (client) -- (api);
\draw[arrow] (api) -- (intent);
\draw[arrow] (intent) -- (agent);
\draw[arrow] (agent) -- (usecases);
\draw[arrow] (agent) -- (eventbus) node[edge_label, above] {Events};
\draw[arrow] (agent) -- (storage) node[edge_label, above, pos=0.6] {Persistence};
\draw[arrow] (agent) -- (llm) node[edge_label, below, pos=0.6] {Model Calls};
\draw[arrow] (eventbus) -- (hitl);
\draw[arrow] (eventbus) -- (storage);
\draw[arrow] (eventbus) -- (obs);
\draw[arrow] (storage) -- (obs);
\draw[arrow] (llm) -- (response);

% --- Manually draw the fill on the background layer (Unchanged) ---
\begin{pgfonlayer}{background}
    \fill[gray!10] (core_box.south west) rectangle (core_box.north east);
    \fill[gray!10] (infra_box.south west) rectangle (infra_box.north east);
\end{pgfonlayer}

\end{tikzpicture}
}
\caption{The ``Maria'' platform's system architecture, illustrating the synergy between the Clean Architecture Core and the Event-Driven Infrastructure.}
\Description{Diagram showing the Maria platform architecture, with a core and infrastructure layer connected asynchronously.}
\label{fig:architecture}
\end{figure*}
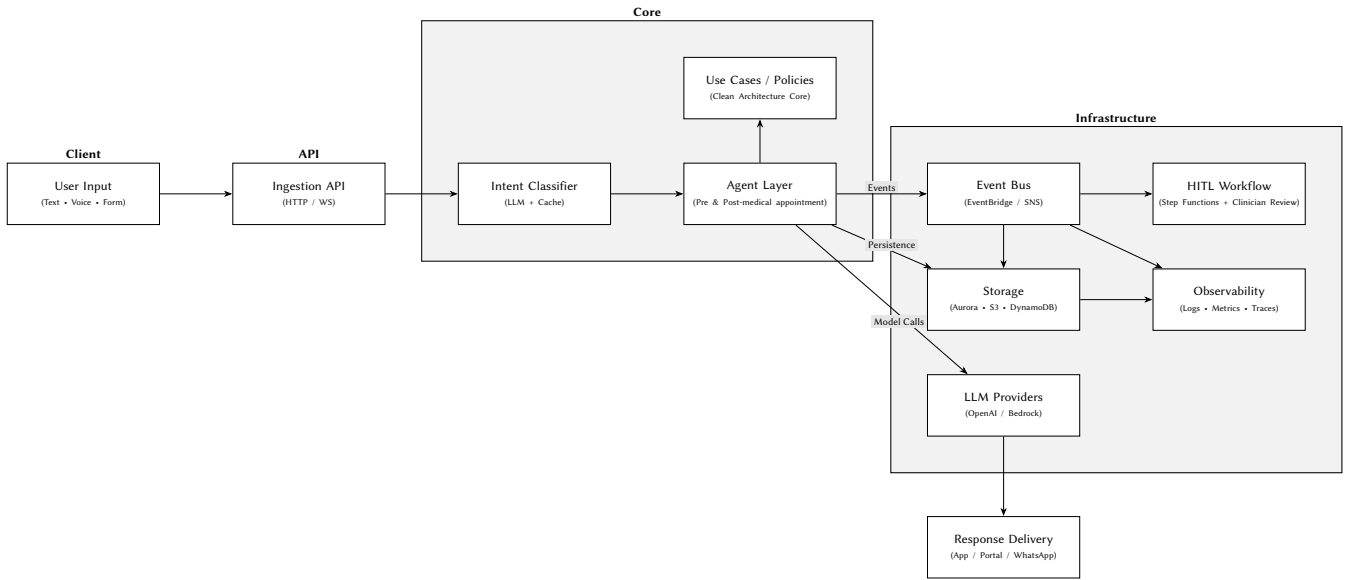

Our first research question (RQ1) asks how Clean Architecture and Event-Driven Architecture (EDA) can be combined to build a resilient and auditable clinical AI system. The design of the ``Maria'' platform achieves this through a clear separation of concerns, as illustrated in Figure \ref{fig:architecture}. The system is logically divided into a technology-agnostic core and a volatile, technology-specific infrastructure layer, which communicate asynchronously.

The \texttt{Core} is designed following the principles of Clean Architecture \cite{10.5555/3175742}. It is the stable center of the system, containing all clinical domain logic and business rules. As data enters from the API, it is first processed by the \texttt{Intent Classifier}. This component, utilizing an LLM with a cache, is responsible for interpreting the initial request and determining the user's intent.

This intent is then passed to the \texttt{Agent Layer}, which contains the application-specific logic that orchestrates the required tasks. It houses our \texttt{Pre-medical appointment} and \texttt{Post-medical appointment} agents. To execute its logic, the \texttt{Agent Layer} depends inward on the \texttt{Use Cases/Policies}, which is the innermost layer containing the stable, high-level clinical policies and business rules (e.g., a post-consultation summary must include a diagnosis). This adherence to the clean architecture dependency rule ensures that the business logic has no dependencies on any outer layer.

Crucially, the entire \texttt{Core} is technology-agnostic. It does not know what database, LLM provider, or messaging system is being used. It communicates its needs to the outer layer via abstract interfaces (ports) for tasks like Persistence, Model Calls, and publishing Events.

The Infrastructure layer contains all concrete, volatile technologies (adapters) that implement the abstract interfaces required by the Core. This layer is designed to be completely replaceable without affecting the Core's business logic. Its components are coordinated by both direct calls and asynchronous events:
\texttt{LLM Providers} implement the large language model calls interface for the Core. When the Agent Layer needs to execute inference, it makes a direct call to this service (e.g., AWS Bedrock \cite{AWS}, OpenAI \cite{OpenAI}). This component is responsible for handling the external API calls and returning the response. As shown in the diagram, it can also route responses to the \texttt{Response Delivery} service.

The \texttt{Storage} component serves two roles. First, it implements the persistence interface, allowing the \texttt{Agent Layer} to make direct, synchronous calls to save or retrieve application state (e.g., using AWS Aurora or DynamoDB\cite{AWS}). Second, it acts as an event consumer, listening to the \texttt{Event Bus} to store immutable event logs or data artifacts (e.g., saving a finalized summary to AWS S3 \cite{AWS}).
    
The \texttt{Event Bus (EDA Core)} is the heart of our asynchronous, decoupled architecture (e.g., AWS EventBridge or SNS \cite{AWS}) \cite{Wang2013, Warrier2025RealTimeAI,Surantha2022}. It implements the events port. When the \texttt{Agent Layer} completes a significant task (like generating a summary), it publishes a domain event. The bus then fans this event out to all subscribed consumers.

The \texttt{HITL Workflow:} A primary event consumer, this component listens for events like \texttt{ClinicalSummaryReadyForReview}. When it receives one, it triggers the formal governance process to queue the AI's output for \texttt{Clinician Review}.

\texttt{Observability} is the service that is the central sink for system health data. It receives signals from two sources: the \texttt{Event Bus} (for business-level events like \texttt{AgentFailed}) and the \texttt{Storage} layer (for data-access logs), aggregating them into \texttt{Logs, Metrics, and Traces}.

This synergistic architecture directly answers RQ1 by establishing a rigorous separation of concerns that supports the strict quality attributes of clinical software.
The adoption of clean architecture ensures maintainability and testability through the dependency rule ; by isolating the clinical domain logic (Core) from volatile infrastructure (e.g., changing LLM providers or database schemas), we can validate complex medical business rules via fast, isolated unit tests without the overhead of external dependencies. Simultaneously, the EDA provides the necessary resilience and auditability for a high-stakes environment.
By decoupling components temporally and spatially \cite{10081336,Surantha2022}, EDA ensures that a failure in a downstream component, such as a latency spike in the HITL workflow, does not propagate to block the synchronous ingestion API, preserving system availability.
Furthermore, the event bus inherently acts as a governance mechanism; every state change is captured as an immutable domain event, providing the granular, end-to-end audit trail required to mitigate the ``responsibility vacuum'' \cite{Anon2025ResponsibilityVacuum} in clinical AI.

This asynchronous, publish-subscribe model \cite{10.1145/857076.857078} is the heart of our EDA. The Event Bus receives this event from the Core and triggers multiple, decoupled downstream consumers in the \texttt{Infrastructure} layer. These consumers include the \texttt{HITL Workflow} (for governance), the \texttt{Observability} pipeline (for MLOps), and \texttt{Response Delivery} (to the client).

\subsection{Agent-Based design and MLOps lifecycle (RQ2)}
\label{sec:system:agents_and_mlops}

Our second research question (RQ2) asks how conceptualizing AI models as autonomous agents, each with a dedicated MLOps lifecycle, improves modularity, governance, and maintainability. Our answer is to shift the fundamental unit of engineering from `the model' to `the agent'. An agent, in our system, is a self-contained component with its own logic, versioned artifacts, and an independent lifecycle, a concept that provides a clear-cut path for modularity and governance.

Unlike a standard microservice, an Agent in our architecture encapsulates not just logic, but probabilistic behavior and memory. Therefore, its lifecycle requires not just unit tests (deterministic), but evaluation suites (probabilistic) that run against versioned of clinical data.

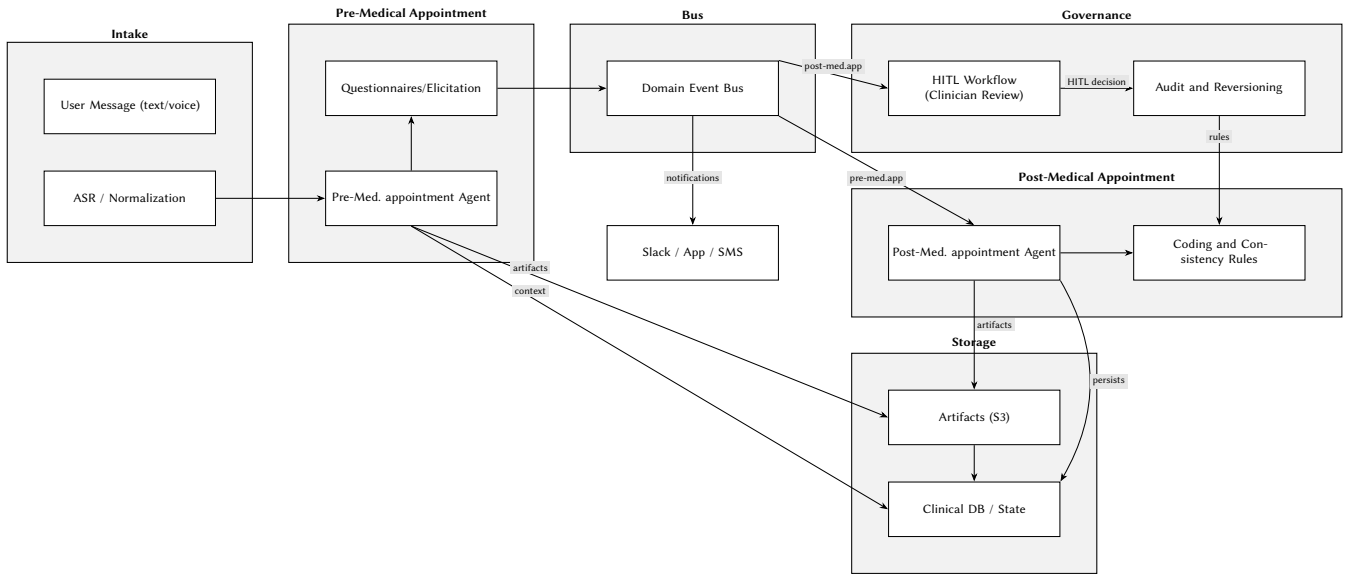
\begin{figure*}[t]
\centering
\resizebox{\textwidth}{!}{%
\begin{tikzpicture}[
    node distance=1cm and 1.5cm,
    % --- Define Styles ---
    comp/.style={
        draw, 
        rectangle, 
        thick,
        fill=white,
        minimum height=1.5cm, 
        text width=4.5cm,
        text centered,
        font=\sffamily\large
    },
    container/.style={
        draw,
        rectangle,
        thick,
        inner sep=1cm,
        label={[font=\sffamily\bfseries\large]north:#1},
        fill=none
    },
    arrow/.style={
        draw, 
        thick, 
        -Stealth
    },
    edge_label/.style={
        fill=gray!20,
        font=\sffamily\small,
        inner sep=2pt,
        pos=0.5
    }
]

% --- Node Placement (Unchanged) ---
\node[comp] (user_message) {User Message (text/voice)};
\node[comp, below=of user_message] (asr) {ASR / Normalization};
\node[comp, right=3cm of asr] (pre_agent) {Pre-Med. appointment Agent};
\node[comp, above=of pre_agent, yshift=0.5cm] (questionnaires) {Questionnaires/Elicitation};
\node[comp, right=3cm of questionnaires] (event_bus) {Domain Event Bus};
\node[comp, right=3cm of event_bus] (hitl) {HITL Workflow (Clinician Review)};
\node[comp, below=3cm of event_bus] (notifications) {Slack / App / SMS};
\node[comp, below=3cm of hitl] (post_agent) {Post-Med. appointment Agent};
\node[comp, right=2cm of hitl] (audit) {Audit and Reversioning};
\node[comp, right=2cm of post_agent] (coding_rules) {Coding and Consistency Rules};
\node[comp, below=of post_agent, yshift=-2cm] (artifacts) {Artifacts (S3)};
\node[comp, below=of artifacts] (clinical_db) {Clinical DB / State};

% --- Fit Containers (Unchanged) ---
\node[container, 
      fit=(user_message) (asr), 
      label={[font=\sffamily\bfseries\large]north:Intake}
      ] (intake_box) {};
\node[container, 
      fit=(pre_agent) (questionnaires), 
      label={[font=\sffamily\bfseries\large]north:Pre-Medical Appointment}
      ] (preconsulta_box) {};
\node[container, 
      fit=(event_bus), 
      label={[font=\sffamily\bfseries\large]north:Bus}
      ] (bus_box) {};
\node[container, 
      fit=(hitl) (audit), 
      label={[font=\sffamily\bfseries\large]north:Governance}
      ] (governance_box) {};
\node[container, 
      fit=(post_agent) (coding_rules), 
      label={[font=\sffamily\bfseries\large]north:Post-Medical Appointment}
      ] (postconsulta_box) {};
\node[container, 
      fit=(artifacts) (clinical_db), 
      label={[font=\sffamily\bfseries\large]north:Storage}
      ] (storage_box) {};

% --- Draw Arrows & Edge Labels ---
\draw[arrow] (asr) -- (pre_agent);
\draw[arrow] (pre_agent.north) -- (questionnaires.south);
\draw[arrow] (questionnaires.east) -- (event_bus.west);

% Event Bus Fan-out
\draw[arrow] (event_bus.north east) -- (hitl.west) node[edge_label, above] {post-med.app};
\draw[arrow] (event_bus.south) -- (notifications.north) node[edge_label, below] {notifications};
\draw[arrow] (event_bus.south east) -- (post_agent.north) node[edge_label, below] {pre-med.app};

% Governance Flow
\draw[arrow] (hitl.east) -- (audit.west) node[edge_label, above] {HITL decision};

\draw[arrow] (post_agent.east) -- (coding_rules.west);
% Path 1: post_agent -> artifacts (Standard straight line)
\draw[arrow] (post_agent.south) -- (artifacts.north) 
    node[edge_label, right, pos=0.4] {artifacts};

% Path 2: post_agent -> clinical_db (CURVED to avoid crossing Artifacts)
% We start at south east and curve to north east
\draw[arrow] (post_agent.south east) to[out=-60, in=60] 
    node[edge_label, right, pos=0.5] {persists} 
    (clinical_db.north east);

% Storage Flow
\draw[arrow] (artifacts.south) -- (clinical_db.north);

% Tweaked anchors to ensure diagonal lines don't clip the Artifacts box
% Path 1: pre_agent -> artifacts
\draw[arrow] (pre_agent.south) -- (artifacts.west) 
    node[edge_label, above, pos=0.25] {artifacts};
% Path 2: pre_agent -> clinical_db
\draw[arrow] (pre_agent.south) -- (clinical_db.west) 
    node[edge_label, above, pos=0.25] {context};
% --- END FIX ---

\draw[arrow] (audit.south) -- (coding_rules.north) 
    node[edge_label, above, pos=0.25] {rules};

% --- Manually draw the fill on the background layer ---
\begin{pgfonlayer}{background}
    \fill[gray!10] (intake_box.south west) rectangle (intake_box.north east);
    \fill[gray!10] (preconsulta_box.south west) rectangle (preconsulta_box.north east);
    \fill[gray!10] (bus_box.south west) rectangle (bus_box.north east);
    \fill[gray!10] (governance_box.south west) rectangle (governance_box.north east);
    \fill[gray!10] (postconsulta_box.south west) rectangle (postconsulta_box.north east);
    \fill[gray!10] (storage_box.south west) rectangle (storage_box.north east);
\end{pgfonlayer}

\end{tikzpicture}
}
\caption{The ``Maria'' platform's Agent-Based Design, showing the flow of information between the Intake, Pre- and Post-Medical appointments agents.}
\label{fig:agent_design}
\end{figure*}

The ``Maria'' platform's clinical workflow, shown in Figure \ref{fig:agent_design}, is structured around two primary autonomous agents that ``bookend'' the entire clinical encounter: the Pre-Medical Appointment Agent and the Post-Medical Appointment Agent.

The process begins in the \texttt{Intake} service, where \texttt{User Message} input (text, voice or form) is processed by an \texttt{ASR / Normalization} (Automatic Speech Recognition-ASR) component. 

The process begins in the \texttt{Intake} service. Here, \texttt{User Message} input (text or voice) is processed by the \texttt{ASR / Normalization} component. This crucial preprocessing step first uses Automatic Speech Recognition (ASR) to transcribe any voice input into raw text. Then, this text, or any direct text input is fed into a normalization service. This service cleans and standardizes the input by removing punctuation, correcting spelling, and converting to lowercase, ensuring that the downstream \texttt{Intent Classifier} receives a clean and predictable machine-readable string.

This data is passed to the \texttt{Pre-Med. Appointment Agent}, which is responsible for the initial patient anamnesis, managing this dialogue via its \texttt{Questionnaires/Elicitation} logic. This agent operates autonomously, as shown by its direct, asynchronous connections to the \texttt{Storage} layer. It persists \texttt{artifacts} (e.g., transcripts) to \texttt{Artifacts (S3)} and \texttt{context} (e.g., a preliminary summary) to the \texttt{Clinical DB / State}.

% The Pre-Medical appointment agent is activated by the \texttt{Intake} service. Its responsibility is to conduct the initial patient anamnesis by engaging in an automated dialogue (\texttt{Questionnaires/Elicitation}). It produces two main outputs: raw \texttt{artifacts} (e.g., transcripts) and structured \texttt{context} (e.g., a preliminary summary, saved to the \texttt{Clinical DB}). It concludes its task by publishing a \texttt{pre-med.app} event to the \texttt{Domain Event Bus}.    

% In the same way, the Post-Medical Appointment agent is triggered by a domain event (e.g., consultation finished) from the main clinical system. It retrieves the full consultation data, applies codified \texttt{Coding and  Consistency Rules}, and generates the final, structured summary. This output is then persisted to the \texttt{Clinical DB} and \texttt{Artifacts (S3)}, and an event is published to the bus, which in turn triggers the HITL governance workflow that updates the \texttt{Coding and Consistency Rules}.

Upon completion, the \texttt{Pre-Medical Appointment} agent's work (via the \texttt{Questionnaires/Elicitation} logic) results in a domain event being published to the central \texttt{Domain Event Bus}. This event-driven approach, which follows a publish-subscribe model, decouples all system components. The \texttt{Bus} then routes event notifications to various subscribers based on the event type. As shown in Figure \ref{fig:agent_design}, a \texttt{post-med.app} event triggers the \texttt{HITL Workflow} in the \texttt{Governance} module for \texttt{Clinician Review}. A subsequent \texttt{HITL decision} event from this workflow then triggers the \texttt{Audit and Reversioning} process. Concurrently, the \texttt{Bus} routes other events, such as \texttt{pre-med.app}, to activate the \texttt{Post-Medical Appointment} agent and sends general \texttt{notifications} to services like \texttt{Slack / App / SMS}. Once activated by its event, the \texttt{Post-Med. appointment Agent} applies its internal logic by referencing the \texttt{Coding and Consistency Rules}, which are themselves updated by the \texttt{Audit and Reversioning} process via the \texttt{rules} path. 

Finally, the agent persists its data, sending \texttt{artifacts} and \texttt{persists} data to the \texttt{Storage} layer. This agent-based design provides high modularity, as each agent's logic is fully encapsulated and communicates with the rest of the system via asynchronous events.

A core tenet of our solution is that if an agent is an autonomous unit of design, it must also be an autonomous unit of deployment and maintenance. Each agent (Pre- and Post-Medical Appointment) has its own dedicated MLOps lifecycle, ensuring that changes to one agent do not affect the other. This ``MedMLOps'' approach is critical for safety and compliance in a regulated environment \cite{Anon2025MedMLOps}. The source of truth for any agent is its source and registry, which versions all its components, including its prompts, behavioral policies, output schemas, and the specific \texttt{Model IDs} it relies on. A change in this registry automatically triggers a CI/CD pipeline. 
Our Source and Registry is implemented using Git and an MLflow registry. The CI/CD pipeline is orchestrated with Github Actions. This pipeline performs some pre-merge checks, such as linting, schema validation, and red-teaming. We enforce a multi-stage privacy protocol: automated Personal Identifiable Information (PII) scanning is executed both during the CI/CD testing of synthetic data and, most importantly, during runtime execution. This ensures that all patient data is sanitized and anonymized immediately upon ingestion, before any payload is transmitted to external LLM providers.

After passing checks, it is validated with offline evaluations against  datasets (it uses custom Python scripts to test against our ``golden set'' stored in S3). Monitoring and Guardrails are implemented using AWS CloudWatch \cite{AWS} Alarms and Sentry.

On success, changes are deployed cautiously, first to a staging deploy using shadow or canary flags, and then to a production deploy with gradual rollouts and auto-rollback mechanisms. Once in production, the observability service becomes critical. It feeds two feedback loops. An automated loop uses monitoring and guardrails to check for data drift or policy violations. A human loop, central to our governance, captures HITL feedback from the clinician review process, which is then used to curate a new ``golden set'' of validated examples. Both feedback loops inform the continuous improvement process, leading to re-scoping, re-tuning, or the creation of a new version, which is then checked back into the registry,  starting the entire cycle anew. This approach directly answers RQ2 by showing how the agent concept provides both structural modularity and the operational maintainability required for a governable AI system.

\subsection{Integrated governance and the Human-in-the-Loop (RQ3)}
\label{sec:system:governance}

Our final pillar, governance, directly answers RQ3, which asks how a Human-in-the-Loop (HITL) model can be technically integrated to ensure safety and accountability. Given that fully autonomous systems are often inappropriate for high-stakes clinical decisions, our architecture treats governance as a first-class, event-driven component, not an afterthought. This approach is designed to fill the ``responsibility vacuum'' \cite{Anon2025ResponsibilityVacuum} by creating a concrete, auditable workflow for every AI-generated output.

\begin{figure*}[t]
\centering
\resizebox{\textwidth}{!}{%
\begin{tikzpicture}[
    node distance=1.5cm and 2.5cm,
    font=\sffamily,
    % --- Define Styles ---
    process/.style={
        draw, thick, 
        rectangle,
        minimum height=1.5cm, 
        text width=4.5cm, 
        text centered,
        fill=white
    },
    storage/.style={
        draw, thick, 
        rectangle,
        minimum height=2cm, 
        text width=4.5cm, 
        text centered,
        fill=gray!10
    },
    arrow/.style={
        draw, 
        thick, 
        -Stealth
    },
    label/.style={
        font=\sffamily\small,
        fill=gray!20,
        inner sep=2pt,
        pos=0.5
    }
]

% --- Node Placement ---
\node[process] (agent_output) {Agent Output \\ \footnotesize (AI Event: completed)};
\node[process, right=of agent_output] (event_bus) {Event Bus };
\node[process, right=of event_bus] (orchestrator) {Validation Orchestrator };
\node[process, right=of orchestrator] (review_ui) {Clinician Review UI \\ \footnotesize (Web Portal / Internal Tool)};
\node[process, above=1.5cm of review_ui, xshift=4cm] (approve) {Version Lock \& Persist \\ \footnotesize (Metadata)};
\node[process, right=3cm of review_ui] (correct) {Feedback Loop \\ \footnotesize (Retraining Dataset / Audit)};
\node[process, below=1.5cm of review_ui, xshift=4cm] (reject) {Rollback \& Alert \\ \footnotesize (Policy Breach / Quarantine)};
\coordinate[right=3cm of correct] (junction);
\node[storage, right=1.5cm of junction] (audit_log) {Audit Record \\ \footnotesize (Immutable Logs + Trace ID + Reviewer ID)};

% --- Draw Arrows & Edge Labels ---
\draw[arrow] (agent_output) -- (event_bus);
\draw[arrow] (event_bus) -- (orchestrator);
\draw[arrow] (orchestrator) -- (review_ui);
\draw[arrow] (review_ui.north) -- (approve.west) node[label, above, sloped] {Approve};
\draw[arrow] (review_ui.east) -- (correct.west) node[label, above] {Correct};
\draw[arrow] (review_ui.south) -- (reject.west) node[label, below, sloped] {Reject};
\draw[arrow] (approve.east) -- (junction);
\draw[arrow] (correct.east) -- (junction);
\draw[arrow] (reject.east) -- (junction);
\draw[arrow] (junction) -- (audit_log);

\end{tikzpicture}
}
\caption{The integrated Human-in-the-Loop (HITL) governance workflow. Every AI-generated event is paused and routed by an orchestrator for explicit human review, which then triggers one of three distinct, fully audited paths: Approve, Correct, or Reject.}
\Description{The integrated Human-in-the-Loop (HITL) governance workflow. Every AI-generated event is paused and routed by an orchestrator for explicit human review, which then triggers one of three distinct, fully audited paths: Approve, Correct, or Reject.}
\label{fig:hitl_workflow}
\end{figure*}
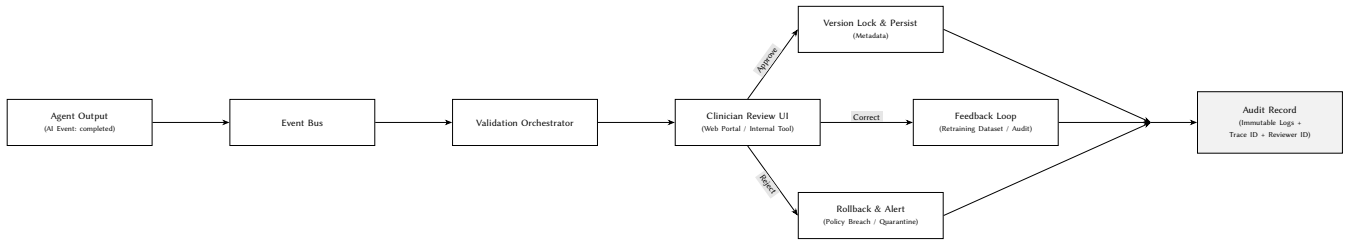

\begin{figure*}[t]
    \centering
    \begin{subfigure}[b]{0.48\textwidth}
        \centering
        % --- IMPORTANT: Rename 'Untitled design(1).png' to 'hitl_clinician_view.png' ---
        \includegraphics[width=\textwidth]{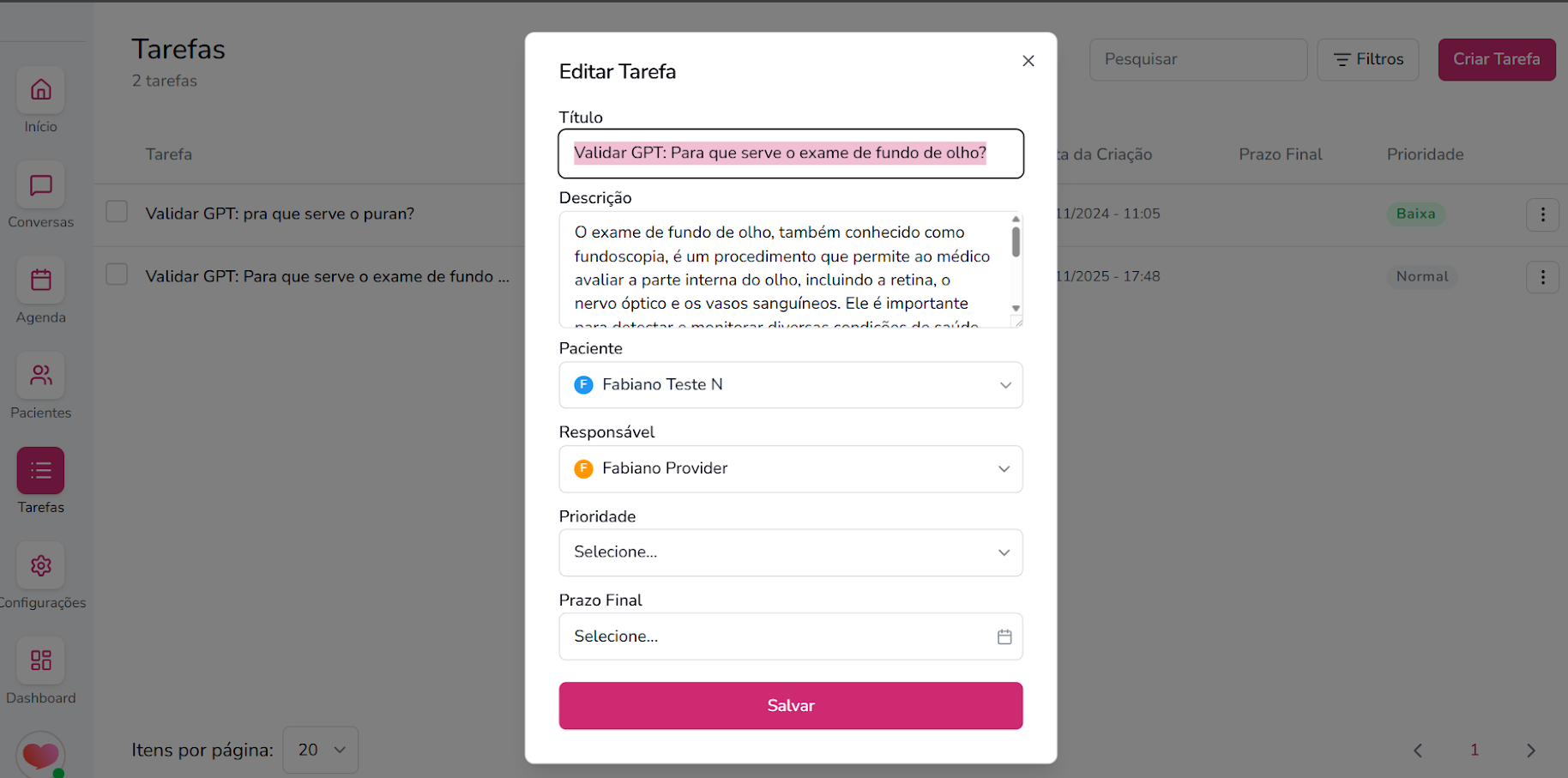}
        \caption{Clinician's Review UI (in Portuguese), showing the AI-generated answer in a task queue for validation.}
        \label{fig:hitl_clinician}
    \end{subfigure}
    \hfill %
    \begin{subfigure}[b]{0.48\textwidth}
        \centering
        % --- IMPORTANT: Rename 'Untitled design.jpg' to 'hitl_patient_view.jpg' ---
        \includegraphics[width=\textwidth]{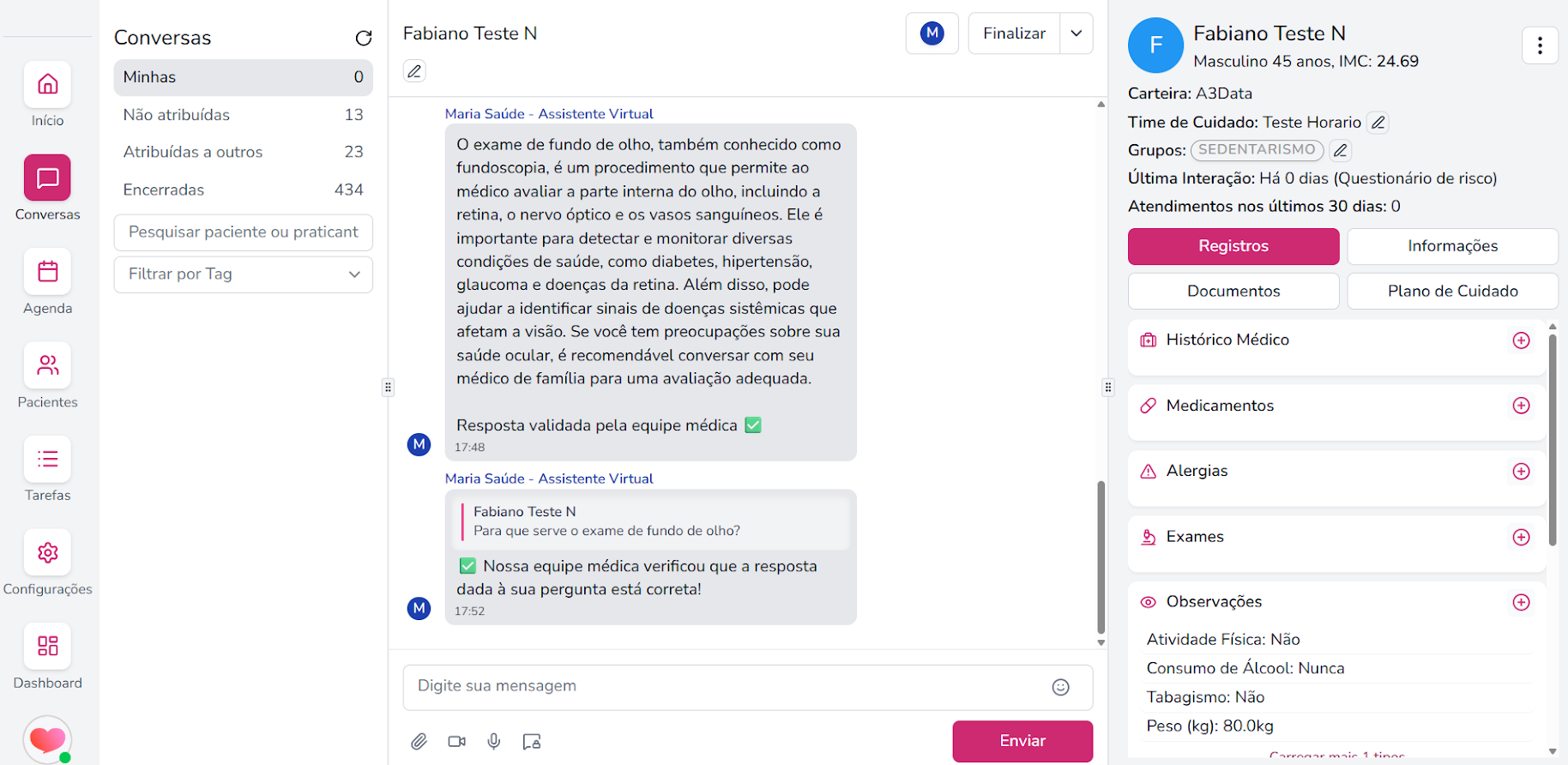}
        \caption{Patient's Chat View (in Portuguese), showing the AI answer followed by the explicit medical team verification.}
        \label{fig:hitl_patient}
    \end{subfigure}
    \caption{Practical application of the HITL workflow in the "Maria" platform. (a) The clinician's validation task. (b) The final, human-curated message delivered to the patient.}
    \Description{Practical application of the HITL workflow in the "Maria" platform. (a) The clinician's validation task. (b) The final, human-curated message delivered to the patient.}
    \label{fig:hitl_screenshots}
\end{figure*}

The logic of this HITL workflow is detailed in Figure \ref{fig:hitl_workflow}. The \texttt{Post-Medical Appointment Agent} does not have the authority to finalize and commit data. Instead, it publishes an \texttt{AI Event: completed} which is routed through the \texttt{Event Bus} to a dedicated \texttt{Validation Orchestrator}. This orchestrator programmatically pauses the workflow and routes the AI's output to the \texttt{Clinician Review UI} (it is a Web Portal or internal tool, observe Figure \ref{fig:hitl_clinician}), where an expert clinician explicitly makes a decision. This step ensures that a qualified human remains the final authority for patient care.

As the diagram illustrates, the clinician's decision defines the subsequent path. This is not a simple pass/fail gate, but a branching logic that ensures safety and continuous improvement.
\begin{itemize}
    \item If the output is \textbf{Approved}, the system proceeds to \texttt{Version Lock \& Persist}. This action ensures that the approved, human-validated output is immutably stored and versioned in the system's production database.
    \item If the output requires \textbf{Correction}, it triggers a \texttt{Feedback Loop}. This path is vital for continuous improvement, as the corrected data is used to enrich a retraining dataset (a ``golden set'') and feeds directly into the MLOps \texttt{Audit} process.
    \item If the output is \textbf{Rejected}, it triggers an immediate \texttt{Rollback \& Alert} mechanism. This could indicate a policy breach or a serious error, leading to the output being quarantined and appropriate stakeholders being notified.
\end{itemize}

Figure \ref{fig:hitl_screenshots} provides a practical example of this workflow from the ``Maria'' platform's production interface (which is in Portuguese). On the left, a clinician sees a task in their ``Tarefas'' (Tasks) queue: ``Validar GPT: para que serve o exame de fundo de olho?'' (Validate GPT: what is the fundus exam for?). This screen, Figure \ref{fig:hitl_clinician} is the implementation of the \texttt{Clinician Review UI} from our flowchart. The clinician can approve or edit, correcting the AI's answer. On the right, the patient's conversation view is shown after the clinician has approved the message. The patient sees not only the AI's answer but also a crucial, human-curated confirmation: ``Nossa equipe médica verificou que a resposta dada à sua pergunta está correta!'' (Our medical team has verified that the answer given to your question is correct!). This explicit confirmation closes the loop, building patient trust and codifying the human's role as the final authority.

Crucially, all three paths: Approve, Correct, and Reject, that conclude by writing to a central \texttt{Audit Record}. This immutable log, which includes trace IDs, reviewer IDs, and timestamps, provides complete end-to-end traceability for every AI-generated output and the human decision that governed it. 

This technical integration is the key to answering RQ3. The HITL workflow is an event-driven component of the MLOps lifecycle itself, providing the essential context, traceability, and feedback loop that make scalable, accountable, and safe clinical AI possible.

% --- NEW FIGURE FOR SCREENSHOTS ---

\section{Results and Discussion: Lessons learned from the field} \label{sec:discussion}

The ``Maria'' platform has been successfully implemented, validating our central hypothesis that a holistic, engineering-first approach is essential for building trustworthy clinical AI. Our results are not measured in model accuracy scores alone, but in the engineering and operational outcomes of our four pillars: maintainability, resilience, modularity, and governance. This section discusses the practical lessons learned from applying these principles in a production environment, offering takeaways for other practitioners.

\subsection{Lesson 1: Treat Clean Architecture as a prerequisite for MLOps}
Our work on RQ1 confirmed that the synergy between Clean Architecture and EDA provides a resilient and auditable system. The hard-earned lesson, however, is that Clean architecture is a foundational enabler for effective MLOps. By adhering to the Dependency Rule, our \texttt{Agent Layer} Figure \ref{fig:architecture} is completely decoupled from the \texttt{LLM Providers}. The Core logic simply makes a call to an abstract interface.

This separation is a critical engineering benefit. It means that A/B testing a new model, switching from OpenAI to Bedrock, for example, or updating a model version is an infrastructure-level change, not a rewrite of core application logic. This isolation makes continuous deployment and model versioning (the pillars of MLOps) practical and safe, rather than a high-risk refactoring exercise.

Empirical validation of this architectural synergy is provided by our production performance metrics. Our AWS CloudWatch \cite{AWS} dashboards for the production API show that the user-facing API gateway maintains an extremely low median latency of 6.13ms and a percentile 90 of 7.69ms.

This high performance is a direct benefit of our EDA. The same dashboard confirms this: while the user receives a near-instant API response, the more complex, asynchronous agent processes are handled in the background. These event-driven tasks complete with a healthy median latency of approximately 130-450ms. The EDA decouples the user's synchronous experience from the agent's asynchronous work, allowing both to be optimized.

Regarding reliability, our Sentry dashboard shows a {0.0\% failure rate} for critical agent endpoints. This is mirrored by our service dashboard, which shows sustained periods with zero HTTP 5xx (server) errors, proving the system is robust and fulfilling its operational requirements. These metrics confirm that the proposed architecture successfully bridges the gap between prototype and production, providing a stable foundation for clinical workflows.

\subsection{Lesson 2: Use EDA as the backbone for auditability and governance}
The EDA pattern was not only crucial for resilience but also proved to be the technical backbone of our entire governance strategy. In high-stakes domains, traceability is non-negotiable, as in clinical systems engineering. Our key insight is that an event-driven system is an auditable system by default.

Every significant action taken by an agent or by a human publishes an immutable event to the \texttt{Domain Event Bus}, see Figure \ref{fig:agent_design}. This event log is the audit trail. When a clinician approves an AI-generated answer (Figure \ref{fig:hitl_clinician}), that ``Approve'' action is an event. When a patient receives the verified message (Figure \ref{fig:hitl_patient}), that is also an event. This provides end-to-end traceability, allowing us to reconstruct the entire lifecycle of any AI decision. This directly addresses the ``responsibility vacuum''  by ensuring no action is ever lost.

\subsection{Lesson 3: The Agent is the correct unit of modularity and lifecycle}
Our work on RQ2 demonstrated that the Agent is a more effective unit of modularity than the model. The practical lesson is that an agent's autonomous MLOps lifecycle is operationally non-negotiable, despite its high upfront cost.

Our system treats each agent as a self-contained unit encompassing its prompts, schemas, policies, and model IDs. Each has its own independent CI/CD pipeline, as described in Section \ref{sec:system:agents_and_mlops}. The temptation is to create a single, shared MLOps pipeline for all agents to save on infrastructure overhead. However, this would create tight coupling; a change to a shared prompt-testing framework for the \texttt{Pre-Med. appointment Agent} could inadvertently break the \texttt{Post-Med. appointment Agent}.

The lesson for practitioners is that the high upfront engineering cost of building and maintaining separate, autonomous lifecycles pays for itself in reduced deployment risk, higher velocity, and robust operational maintainability.

\subsection{Lesson 4: HITL is an MLOps data source, not just a safety check}
This was our most significant finding related to RQ3. Most systems treat HITL as a simple pass/fail safety check. Our experience confirms that HITL's primary value is the most critical data source for the MLOps continuous improvement loop.

To provide concrete evidence for this lesson, we analyzed the production audit logs from the first year of the HITL workflow (since October 2024), which comprised 350 total AI-generated validation tasks. Of these, 144 tasks were explicitly actioned by clinicians (the remainder were simple, informational answers that clinicians deemed unnecessary to formally validate). From this set of 144 explicit reviews, our HITL data reveals a clear and powerful pattern that validates our entire architecture:

\begin{itemize} \item {81\% (117 tasks) were ``Approved''}: The clinician validated the AI's response with no modifications. \item {19\% (27 tasks) were ``Corrected''}: The clinician edited the response before approving it. \item 0\% (0 tasks) were ``Rejected'': We have had zero instances of a clinician needing to fully reject an AI answer for a policy breach or for being dangerously incorrect. \end{itemize}

This data provides two critical insights. First, the 0\% rejection rate suggests that our upstream engineering, the autonomous agent lifecycles, pre-merge checks, and guardrails, is effectively preventing catastrophic failures before they reach the generation phase. The ``Rollback and Alert'' mechanism serves as an essential, but thus far, unneeded, safety net. 

Second, the 19\% ``Corrected'' cohort is the most valuable output of our system. These 27 corrections are not failures; they are high-quality, expert-curated data points that form the core of our ``golden set'' for the MLOps \texttt{Offline Eval}. This shows that the HITL workflow is functioning exactly as designed: as both a final safety net and a powerful, self-sustaining data engine for continuous improvement.

% The governance workflow, depicted in Figure \ref{fig:hitl_workflow}, is not just a UI; it is a core MLOps feature. The ``Approve'' path is a confirmation, but the ``Correct'' path is the most valuable. When a clinician edits an AI's output, they are not logging a failure; they are providing expert-curated, high-quality training data for improving the user's experience. This \texttt{HITL Feedback} is captured, routed to our \texttt{Audit Record}, and used to build a new golden set for the \texttt{Offline Eval}.

This closes the loop, technically integrating the human's expertise back into the automated pipeline and ensuring the agent continuously improves from its real-world performance.

\subsection{Lesson 5: Explicit Governance Builds Both Patient and Clinician Trust}

A final, crucial lesson is socio-technical. Building a governable system is only half the battle; the other half is communicating that governance. This lesson is not just a best practice but also a direct alignment with emerging legal frameworks like Brazil's Bill 2338/2023  \cite{senado:pl2338}, which mandates principles of transparency, explainability, intelligibility, and auditability.

Our HITL process serves as a practical implementation of these principles. For clinicians, the \texttt{Clinician Review UI} (Figure \ref{fig:hitl_clinician}) provides {explainability} by default; they are not forced to accept a black-box output but are presented with the AI's reasoning for review. This trust was captured in direct qualitative feedback. After an engineering update improved the AI's guardrails for complex queries, a clinical leader shared a screenshot of the new AI response (which safely deferred to the human team for a question about medication and anxiety), commenting:

\begin{quote}
    ``Everyone, have those changes to the AI been made? Because look at this amazing answer!''
\end{quote}

This feedback is crucial: the clinician's confidence came precisely because the AI demonstrated its safety, correctly escalated to the human team, and operated as a true assistant, not an autonomous decision-maker. Their subsequent action (Approve, Correct, Reject) is then captured by our \texttt{Audit Record} (Figure \ref{fig:hitl_workflow}), directly fulfilling the principle of {auditability}.

% Our HITL process serves as a practical implementation of these principles. For clinicians, the \texttt{Clinician Review UI} (Figure \ref{fig:hitl_clinician}) provides {explainability} by default; they are not forced to accept a black-box output but are presented with the AI's reasoning for review. Their subsequent action (Approve, Correct, Reject) is then captured by our \texttt{Audit Record} (Figure \ref{fig:hitl_workflow}), directly fulfilling the principle of {auditability}.

For patients, the lesson is even more direct. The regulation requires {transparency} (informing users they are interacting with AI) and {intelligibility} (using clear language). As shown in Figure \ref{fig:hitl_patient}, our system first delivers the AI's answer, clearly sourced from the virtual assistant. More importantly, after the HITL review, we explicitly state: Our medical team has verified that the answer... is correct!. This message fulfills the legal and ethical need for transparency, manages patient expectations by clarifying the AI's role as an assistant, and builds trust by proving that a human expert clinician remains the ultimate authority. This built-in, demonstrable governance is no longer just good engineering; it is a critical requirement for deploying AI systems in regulated markets.

\section{Conclusion and Future work } 
\label{sec:conclusion}

The successful integration of AI into clinical settings is fundamentally an engineering challenge, not just one of data science. The field has been hampered by monolithic prototypes that lack the reliability, maintainability, and governance required for high-stakes healthcare environments. This paper presented ``Maria'', an industrial case study of a production-grade clinical AI platform that addresses this challenge through a holistic, engineering-first approach.

Our primary contribution is a reference architecture for building trustworthy AI systems. We have demonstrated how a synergistic combination of \textbf{Clean Architecture} and \textbf{Event-Driven Architecture (EDA)} provides resilience, testability, and inherent auditability (answering RQ1). We showed how conceptualizing AI as autonomous \textbf{Agents}, each with its own dedicated \textbf{MLOps lifecycle} (Figure \ref{fig:agent_design}), is critical for modularity, maintenance, and managing complexity (answering RQ2). Most importantly, we presented how a \textbf{Human-in-the-Loop (HITL)} workflow is not an add-on but a core, event-driven component of the MLOps lifecycle itself (Figure \ref{fig:hitl_workflow}). This integration provides the technical foundation for accountability, fills the ``responsibility vacuum'', and ensures our system aligns with emerging regulations (answering RQ3).

Future work will focus on expanding the  ``Maria'' platform's capabilities to further automate and augment clinical workflows. The immediate roadmap involves developing specialized agents to support various clinical specialties, such as cardiology, orthopedics, or mental health. 

Concurrently, we plan to conduct a broader empirical benchmarking study across these diverse specialties. This initiative aims to rigorously test the generalizability of our architectural and governance patterns beyond primary care, ensuring they remain robust under varying clinical data densities and workflow complexities. This presents a significant MLOps and operational challenge: evolving our single-agent lifecycle, Section \ref{sec:system:agents_and_mlops}, into a scalable framework capable of managing dozens of specialized agents in parallel. Beyond the complexity of managing unique prompts, policies, and HITL feedback loops for each specialty, we anticipate significant hurdles in cost management and state orchestration. Scaling to multiple concurrent agents increases token usage, requiring rigorous cost-control strategies at the infrastructure layer.

Throughout this evolution, our guiding principle remains unchanged: \textbf{the clinician stays at the center}. Our goal is not to replace clinical judgment, but build a powerful assistant that removes administrative burdens. All future automation will be built upon the governance-first framework presented. This means enhancing the HITL workflow (Figure \ref{fig:hitl_screenshots}) as a tool, creating more sophisticated feedback loops where the system learns the preferences of individual specialists, and ensuring that ``Maria'' evolves as a trusted, governable, and indispensable augmentation to clinical practice.

% \section{Acknowledgments}

\bibliographystyle{ACM-Reference-Format}
\bibliography{software}

\end{document}